%% file: main.tex
\documentclass[runningheads]{llncs}

\usepackage{eccv}

\usepackage{eccvabbrv}

\usepackage{graphicx}
\usepackage{booktabs}
\usepackage{multirow}
\usepackage{amssymb}
\usepackage{xcolor}
\usepackage{pifont} % 提供更美观的勾叉符号
\usepackage[accsupp]{axessibility}  % Improves PDF readability for those with disabilities.

\usepackage{hyperref}

\usepackage{orcidlink}

\begin{document}

% ---------------------------------------------------------------
% TODO REVIEW: Replace with your title
\title{NeSy-Route: A Neuro-Symbolic Benchmark for Constrained Route Planning in Remote Sensing} 

% TODO REVIEW: If the paper title is too long for the running head, you can set
% an abbreviated paper title here. If not, comment out.
\titlerunning{NeSy-Route: Neuro-Symbolic Route Planning Benchmark}

% TODO FINAL: Replace with your author list. 
% Include the authors' OCRID for the camera-ready version, if at all possible.
% \author{Ming Yang\inst{1}\orcidlink{0000-1111-2222-3333} \and
% Zhi Zhou\inst{2,3}\orcidlink{1111-2222-3333-4444} \and
% Shi-Yu Tian\inst{3}\orcidlink{2222--3333-4444-5555} \and Kun-Yang Yu\inst{3}\orcidlink{2222--3333-4444-5555}\and Lan-Zhe Guo\inst{3}\orcidlink{2222--3333-4444-5555}\and Yu-Feng Li\inst{3}\orcidlink{2222--3333-4444-5555}}

\author{Ming Yang\inst{1,2} \and
Zhi Zhou\inst{1}\textsuperscript{*}\and
Shi-Yu Tian\inst{1,2} \and Kun-Yang Yu\inst{1,2}\and Lan-Zhe Guo\inst{1,3}\and Yu-Feng Li\inst{1,2}\thanks{Corresponding authors.}}

% \author{Ming Yang\inst{1,2} \and
% Zhi Zhou\inst{1}\and
% Shi-Yu Tian\inst{1,2} \and Kun-Yang Yu\inst{1,2}\and
% Lan-Zhe Guo\inst{1,3}\and
% Yu-Feng Li\inst{1,2}\thanks{Corresponding author.}}

% TODO FINAL: Replace with an abbreviated list of authors.
\authorrunning{M. Yang et al.}
% First names are abbreviated in the running head.
% If there are more than two authors, 'et al.' is used.

% TODO FINAL: Replace with your institution list.
\institute{National Key Laboratory for Novel Software Technology, Nanjing University, China \and
School of Artificial Intelligence, Nanjing University, China\\ \and School of Intelligence Science and Technology, Nanjing University, China
% \email{\{yangm\}@lamda.nju.edu.cn}}
\email{\{yangm,zhouz,tiansy,yuky\}@lamda.nju.edu.cn, \\ \{guolz,liyf\}@nju.edu.cn}}
\maketitle
\begin{abstract}
  % Existing remote sensing benchmarks for Multimodal Large Language Models (MLLMs) primarily evaluate perception and reasoning tasks but overlook planning capability, which is vital for real-world applications like disaster relief and ecological field surveys. To bridge this gap, we introduce NeSy-Route, a large-scale neuro-symbolic benchmark for constrained route planning in remote sensing. NeSy-Route assesses performance through three hierarchical tasks including textual constraint understanding, text to image constraint alignment, and constrained route planning. By integrating high-fidelity semantic masks with heuristic search, NeSy-Route provide 10,821 route samples with mathematically proven global optimality. Evaluations of state-of-the-art MLLMs reveal a significant bottleneck in translating symbolic rules into precise spatial execution. By offering a rigorous diagnostic framework to identify cognitive failures, NeSy-Route fills a critical research gap and steers reasoning intelligence toward decision-making intelligence.
  % RS 很重要且同时包含感知、推理与规划
  % 现有评估数据集无法评估推理，要么因为数据难收集，要么因为评估协议受限。
  Remote sensing underpins crucial applications such as disaster relief and ecological field surveys, where systems must understand complex scenes and constraints and make reliable decisions. Current remote-sensing benchmarks mainly focus on evaluating perception and reasoning capabilities of multimodal large language models (MLLMs). They fail to assess planning capability, stemming either from the difficulty of curating and validating planning tasks at scale or from evaluation protocols that are inaccurate and inadequate.
  % % 我们提出一套可扩展的神经符号评估框架：合成可验证数据解决可扩展问题；多层次神经符号评估解决评估不准确不充分问题。
  % To address these limitations, we introduce NeSy-Route, benchmarking constrained route planning in remote sensing. Within this, we propose an automated data-generation pipeline that combines high-fidelity semantic masks with heuristic search to create diverse route-planning tasks with verified optimal solutions. Furthermore, a three-level hierarchical neuro-symbolic evaluation protocol is developed to enable accurate assessment of planning performance and supports fine-grained analysis.
  % % 数据集很大；评估很详细；再给出一些结论
  % Based on this framework, we release the NeSy-Route dataset comprising 10,821 route-planning tasks, which is nearly 10 times larger than the largest prior benchmark in assessing planning. We conduct a comprehensive evaluation of various state-of-the-art closed-source and open-source MLLMs, using eight metrics to simultaneously assess perception, reasoning, and planning capabilities for the first time.
  % Our study demonstrates XXX observations: XXX; XXX; XXX
  To address these limitations, we introduce NeSy-Route, a large-scale neuro-symbolic benchmark for constrained route planning in remote sensing. 
  Within this benchmark, we introduce an automated data-generation framework that integrates high-fidelity semantic masks with heuristic search to produce diverse route-planning tasks with provably optimal solutions. This allows NeSy-Route to comprehensively evaluate planning across 10,821 route-planning samples, nearly 10 times larger than the largest prior benchmark. Furthermore, a three-level hierarchical neuro-symbolic evaluation protocol is developed to enable accurate assessment and support fine-grained analysis on perception, reasoning, and planning simultaneously. Our comprehensive evaluation of various state-of-the-art MLLMs demonstrates that existing MLLMs show significant deficiencies in perception and planning capabilities. We hope NeSy-Route can support further research and development of more powerful MLLMs for remote sensing.% 现有模型在规划任务上不足，感知和规划能力都不足；% 社区影响
  The dataset and code are available at \url{https://mingyang1010.github.io/NeSy-Route/}.
\end{abstract}

\input{sections/1.Introduction}
\input{sections/2.Related_work}
\input{sections/3.Data_Generation}
\input{sections/4.Benchmark}
\input{sections/5.Experiment}
\input{sections/6.Conclusion}
\input{sections/7.Acknowledgements}
\bibliographystyle{splncs04}
\bibliography{main}
\end{document}

%% file: sections/1.Introduction.tex
\section{Introduction}
In recent years, multimodal large language models (MLLMs)~\cite{gemini3pro2025,bai2025qwen3,qwen3.5,LLaVA-OneVision-1.5,singh2025openai} have achieved remarkable progress in visual perception~\cite{zhou2023ods,zhou2024decoop, lv2025bmip}  and complex reasoning~\cite{liu2026zoomearth,tian2025tabularmath,zhou2026theoretical,yu2026thinking, yang2025neuro, shao2025chinatravel}. Remote sensing imagery, a crucial source for observing the Earth's surface, supports applications such as environmental monitoring~\cite{marsocci2024pangaea,li2019coverage}, disaster assessment~\cite{dell2012remote}, agricultural management~\cite{zhang2016crop}, and urban planning~\cite{zhu2019understanding}. Among these applications, route planning is important for tasks such as emergency response and resource allocation. However, conventional route-planning systems rely on costly terrestrial surveying~\cite{hu2024towards} to construct road maps, making them vulnerable in disaster-stricken or poorly mapped regions.

MLLMs offer a potential alternative by processing remote sensing imagery for route planning, yet existing remote-sensing benchmarks~\cite{wang2025xlrs,li2024vrsbench} mainly evaluate perception and reasoning, providing little insight into constrained planning abilities, which stem either from the difficulty of curating and validating planning tasks at scale or from evaluation protocols that are inaccurate and inadequate.

\begin{figure}[t!]
  \centering
  \includegraphics[width=\linewidth]{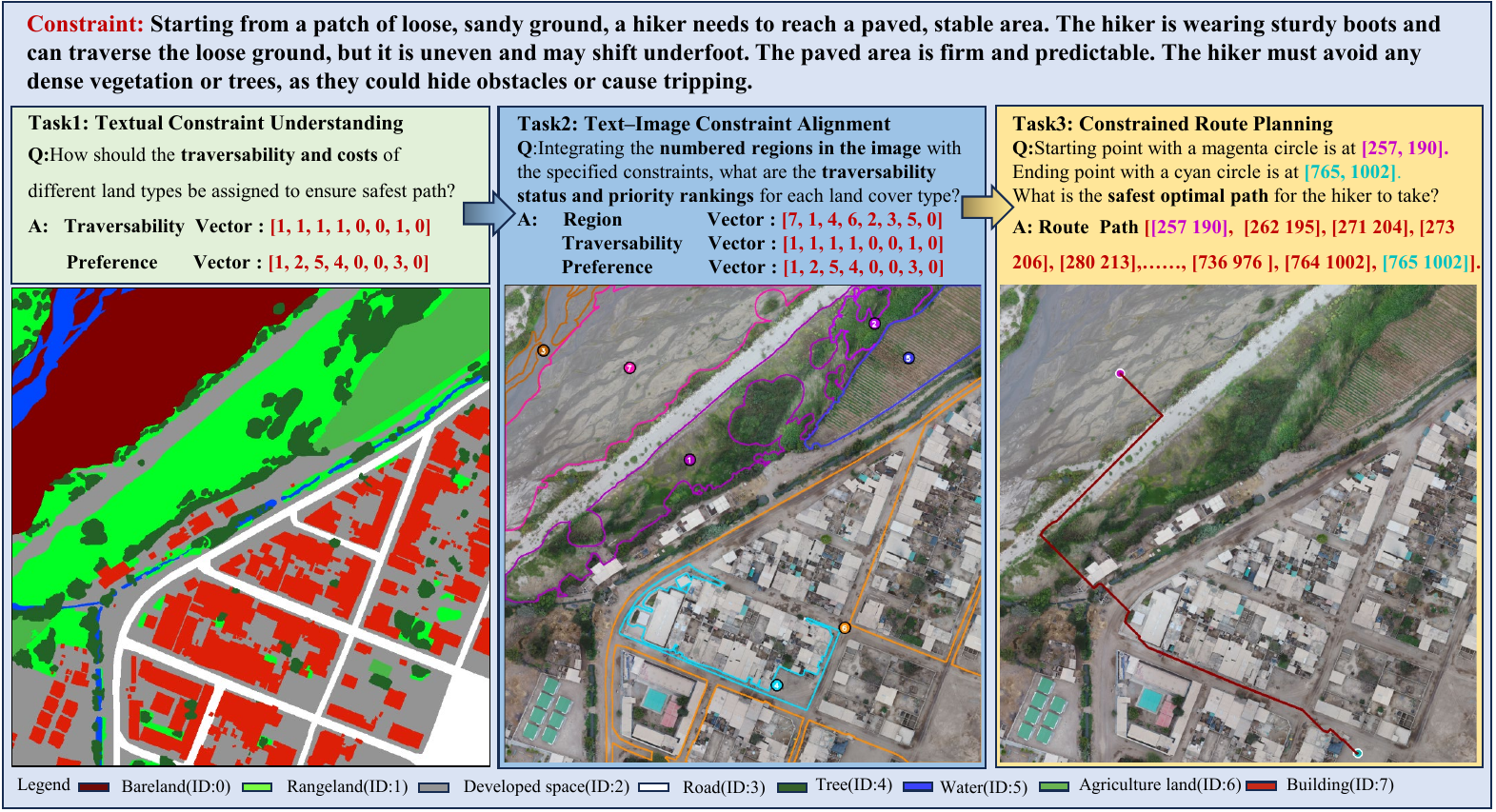}
  \caption{A typical example from NeSy-Route Benchmark. NeSy-Route evaluates MLLMs through a hierarchical reasoning pipeline consisting of three integrated tasks. Task 1 involves extracting symbolic traversability and cost vectors from the provided scenario. Task 2 focuses on anchoring these symbolic constraints to specific identified regions within the remote sensing image. Task 3 assesses the capability to generate a sparse waypoint trajectory that avoids obstacles and minimizes the cumulative cost.}
  \label{fig:behcmark}
\end{figure}

To address these challenges, we introduce NeSy-Route, a neuro-symbolic evaluation benchmark for constrained route planning built on OpenEarthMap~\cite{xia2023openearthmap}, designed to hierarchically assess route planning capability of MLLMs in remote sensing.  The benchmark organizes its task across three levels:
\begin{enumerate}
    \item \textbf{Textual Constraint Understanding Task.}This task evaluates the capacity of MLLMs to decode natural language instructions into formal symbolic logic through 3,607 problems, establishing a critical prerequisite for performing constrained route planning.
    \item \textbf{Text–Image Constraint Alignment Task.} This task challenges models to anchor textual constraints onto 12,975 samples with identified visual regions, requiring joint reasoning to determine terrain-specific traversability and priority rankings.
    \item \textbf{Constrained Route Planning Task.} This task assesses end-to-end planning solver performance by requiring the generation of executable waypoint sequences between designated start and end points that strictly honor both topological barriers and land type constraints across 10,821 samples.
\end{enumerate}
In summary, our main contributions are as follows:
\begin{enumerate}
    \item We propose NeSy-Route, the first neuro-symbolic evaluation benchmark for route planning in remote sensing. It hierarchically assesses the constrained planning capabilities of MLLMs with a symbolic evaluator across three dimensions: textual constraint understanding,  text–image constraint alignment, and constrained route planning.
    \item We design an automated, symbolized data generation framework and develop a symbolized route planning evaluator, establishing a closed-loop framework that integrates neuro generation with symbolic verification.
    \item We conduct a comprehensive evaluation of state-of-the-art MLLMs and reveal significant limitations in their perception, reasoning, planning capabilities simultaneously in real remote sensing scenarios. We further discuss potential directions for MLLMs in advancing planning capabilities.
\end{enumerate}

%% file: sections/2.Related_work.tex
\section{Related Work}

\subsection{Remote Sensing Benchmark}
In recent years, the rapid advancement of MLLMs has accelerated the transition of remote sensing research from fundamental perception tasks to complex reasoning scenarios featuring smaller targets or higher spatial resolutions. While early benchmarks primarily addressed semantic segmentation ~\cite{xia2023openearthmap,wang2021loveda,boguszewski2021landcover}, object detection~\cite{li2020object,lee2025kfgod} and classification~\cite{xia2017aid}, more recent efforts have expanded into VQA task such as spatial relationship reasoning~\cite{li2024vrsbench,luo2406skysensegpt,tian2026last,muhtar2024lhrs}, temporal change detection~\cite{li2024show}, and climate-related risk prediction~\cite{wang2025disasterm3}. However, these datasets are predominantly designed for descriptive reasoning regarding specific objects and their mutual relationships, thereby neglecting the challenges of  route planning within complex remote sensing contexts. XLRS-Bench~\cite{wang2025xlrs}manually annotates 16 sub-tasks and is the first to introduce the Route Planning task in the remote sensing domain, which requires the model to select the correct route description from several candidate options. However, such an evaluation paradigm cannot quantitatively assess planning capability in real remote sensing environments, where routes must be generated through spatial reasoning rather than selected from predefined candidate options.

\subsection{Multimodal Large Language Model}
MLLMs have advanced rapidly in recent years, and most of them now support inputs at native image resolution. Representative closed-source models include Gemini-3-Pro~\cite{gemini3pro2025}, GPT-5.1~\cite{singh2025openai}, and Qwen3-VL-Plus~\cite{bai2025qwen3}, which demonstrate strong capabilities in perception and reasoning. Among open-source models, dense architectures include LLaVA-OneVision~\cite{LLaVA-OneVision-1.5}, the InternVL-3.5-8B~\cite{wang2025internvl3}, Qwen3-VL-8B~\cite{bai2025qwen3}, and Qwen3.5-27B~\cite{qwen3.5}, while mixture-of-experts (MoE) architectures include Qwen3-VL-30B-A3B~\cite{bai2025qwen3} and Qwen3-VL-235B. These open-source models also achieve competitive performance.MLLMs have also shown promising progress in the remote sensing domain. Geochat~\cite{kuckreja2024geochat}, built on the LLaVA-1.5 architecture~\cite{liu2024improved}, is trained to enable multi-task conversational capabilities. SkySenseGPT~\cite{Guo_2024_CVPR_skysense}, also based on LLaVA-1.5, achieves improved performance across a variety of remote sensing tasks through large-scale training. EarthGPT~\cite{zhang2024earthgpt} focuses on understanding of multi-sensor remote sensing data.

%% file: sections/3.Data_Generation.tex
\section{Automated Data Generation Framework}
\begin{figure}[t!]
  \centering
  \includegraphics[width=\linewidth]{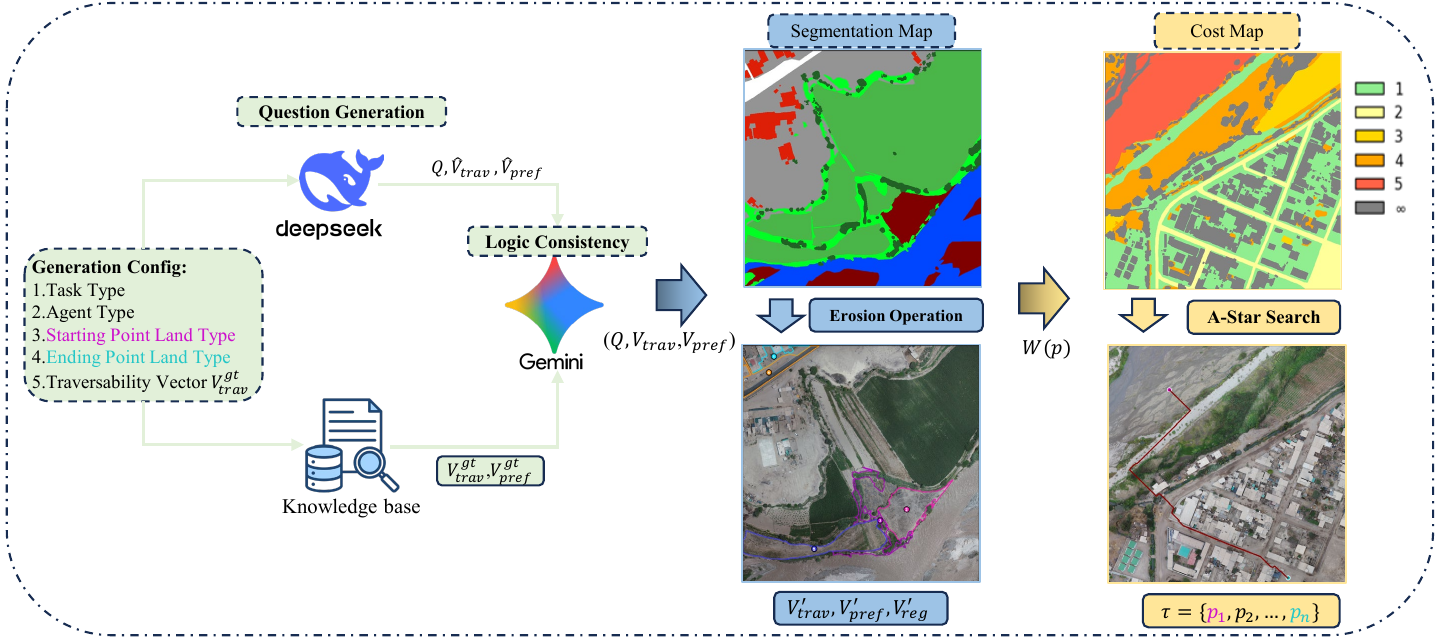}
  \caption{An overview of automated data generation framework. The pipeline integrates dual-LLM logical verification for query synthesis, morphological erosion for semantic visual grounding, and constrained A-Star search algorithm for deriving mathematically optimal trajectories under symbolic rules.}
  \label{fig:datageneration}
\end{figure}

As illustrated in \cref{fig:datageneration}, the automated framework generates image-text pairs, GT vectors, and optimal trajectories for sub-tasks through a three-stage process.
\subsection{Knowledge Base Construction}
\label{sec:kb_cons}
Following OpenEarthMap~\cite{xia2023openearthmap} standards, we define our knowledge base(KB) constraints across eight land-cover types: Bareland (ID:0), Rangeland (ID:1), Developed space (ID:2), Road (ID:3), Tree (ID:4), Water (ID:5), Agriculture land (ID:6), and Building (ID:7).  Let $\mathcal{C} = \{c_0, c_1, \dots, c_{L-1}\}$ be the set of $L=8$ predefined land-cover classes. Specifically, we define three levels of traversability: always traversable, conditionally traversable, and non-traversable:
\begin{itemize}
\item \textbf{Always Traversable ($\mathcal{T}_A$)}: Terrains that are always passable by default. These areas are considered accessible unless explicitly restricted by specific problem constraints.
\item \textbf{Conditionally Traversable ($\mathcal{T}_C$)}: Terrains that are initially impassable. These areas are considered blocked unless specific conditions for passage are met, as defined by the problem constraints.
\item \textbf{Non-traversable ($\mathcal{T}_N$)}: Terrains that are permanently impassable. These areas are permanent obstacles that cannot be crossed under any circumstances, regardless of task-specific requirements.
\end{itemize}

\begin{table}[tbp]
\centering
\caption{Agent-Specific Traversability Definitions}
\label{tab:agent_traversability}
\small
\begin{tabular}{l p{3cm} p{3cm} p{3cm}}
\toprule
\textbf{Agent} & \textbf{$\mathcal{T}_A$} & \textbf{$\mathcal{T}_C$} & \textbf{$\mathcal{T}_N$} \\ \midrule
Pedestrian & \{$c_2, c_3$\} & \{$c_0, c_1, c_4, c_6$\} & \{$c_5, c_7$\} \\
Car        & \{$c_2, c_3$\} & \{$c_0, c_1, c_6$\} & \{$c_4, c_5, c_7$\} \\
Drone      & \{$c_0, c_1, c_2, c_3, c_5, c_6$\} & \{$c_4, c_7$\} & $\emptyset$ \\
Boat       & \{$c_5$\} & $\emptyset$ & \{$c_0, c_1, c_2, c_3, c_4, c_6, c_7$\} \\ \bottomrule
\end{tabular}
\end{table}

Based on the land-cover physical properties, we define four agent types (pedestrian, vehicle, drone, and boat) with traversability constraints in ~\cref{tab:agent_traversability}.

\begin{table}[!ht]
\centering
\caption{Agent-Specific Task Priority Rankings}
\label{tab:agent_priority}
\small
\begin{tabular}{l |p{2.5cm}| p{2.5cm}|p{2.5cm}|p{2.5cm}}
\toprule
\textbf{Agent} & \textbf{Fastest} & \textbf{Comfort} & \textbf{Safest} & \textbf{Shortest} \\ \midrule
Pedestrian & $c_0 = c_1 = c_2 = c_3 = c_4 = c_6$ & $c_2 > c_3 > c_6 > c_1 > c_0 > c_4$ & $c_2 > c_3 > c_6 > c_1 > c_0 > c_4$ & $c_0 = c_1 = c_2 = c_3 = c_4 = c_6$ \\
Car        & $c_3 > c_2 > c_6 > c_1 > c_0$ & $c_3 > c_2 > c_6 > c_1 > c_0$ & $c_3 > c_2 > c_6 > c_1 > c_0$ & $c_0 = c_1 = c_2 = c_3 = c_6$ \\
Drone      & $c_0 = c_1 = c_2 = c_3 = c_5 = c_6 = c_7 > c_4$ & $c_0 = c_1 = c_2 = c_3 = c_6 = c_7 > c_5 > c_4$ & $c_0 = c_1 = c_2 = c_3 = c_6 > c_4 = c_5 = c_7$ & $c_0 = c_1 = c_2 = c_3 = c_5 = c_6 = c_7 > c_4$ \\
Boat       & $c_5$ & $c_5$ & $c_5$ & $c_5$ \\ \bottomrule
\end{tabular}
\end{table}

 Additionally, we define four routing objectives: shortest, fastest, safest, and most comfortable in ~\cref{tab:agent_priority}. For each routing objective, a priority ranking of land cover types is established, which determines the preferred terrains for each agent depending on the specific goal.
% To facilitate neural-symbolic reasoning, we establish a formal taxonomy of land cover types and a hierarchical system for traversability logic. These rules provide the symbolic foundation upon which the vectors $\mathbf{V}_{\text{trav}}$ and $\mathbf{V}_{\text{pref}}$ are constructed.

\subsection{Problem Formulation}
Given a remote sensing visible light image $I \in \mathbb{R}^{H \times W \times 3}$ and a natural language query $Q$ describing a route planning scenario, we formulate the constrained route planning task as a hierarchical evaluation process. The objective is to derive symbolic constraints from $Q$, align them with the visual contexts in $I$, and generate an optimal trajectory $\tau^*$ that minimizes cost function derived from these constraints.

\subsubsection{Task 1: Textual Constraint Understanding}
Given $Q$, the model extracts symbolic traversability and preference vectors via $f_{\text{text}}: Q \to (\mathbf{V}_{\text{trav}}, \mathbf{V}_{\text{pref}})$.
\begin{itemize}
    \item $\mathbf{V}_{\text{trav}} \in \{0, 1\}^L$: Binary vector where $V_{\text{trav}}[i] = 1$ if class $c_i$ is traversable for the specific agent, and $0$ otherwise.
    \item $\mathbf{V}_{\text{pref}} \in \{0, \dots, K\}^L$: Ordinal vector representing the priority of class $c_i$ under the specified task objective, where $K$ is the maximum priority tier. The higher the value, the higher the priority.
\end{itemize}

\subsubsection{Task 2: Text–Image Constraint Alignment}
Given $Q$ and the image $I$ annotated with regions and their corresponding IDs, the model outputs symbolic region, traversability, and preference vectors: 
$f_{\text{align}}: (Q, I) \to (\mathbf{V}'_{\text{reg}}, \mathbf{V}'_{\text{trav}}, \mathbf{V}'_{\text{pref}})$.

\begin{itemize}
    \item $\mathbf{V}'_{\text{reg}} \in \mathbb{N}^L$: A vector where $V_{\text{reg}}[i] = \text{Region\_ID}$ if land cover $c_i$ is present and identified in the image, and $0$ otherwise.
    \item  $\mathbf{V}'_{\text{trav}}$ and $\mathbf{V}'_{\text{pref}}$: Represent the traversability and preference status specifically for the detected regions to evaluate whether the model can correctly apply symbolic rules to the finite set of terrains visible in $I$.
\end{itemize}
Based on $\mathbf{V}'_{\text{pref}} $, we define cost map $W(p)$ for every pixel $p \in I$ corresponding to its land-cover class $c(p)$:
\begin{equation}
\label{eq:costmap}
W(p) = 
\begin{cases} 
\infty, & \text{if } V_{\text{trav}}[c(p)] = 0 \\
\max(\mathbf{V}'_{\text{pref}}) - V_{\text{pref}}[c(p)] + 1, & \text{if } V_{\text{trav}}[c(p)] = 1
\end{cases}
\end{equation}

\subsubsection{Task 3: Constrained Route Planning}
Given $Q$, $I$, start point $S \in \mathbb{N}^2$ and end point $E \in \mathbb{N}^2$, the objective is to generate a trajectory $\tau = \{p_1, p_2, \dots, p_n\}$ that minimizes the cumulative spatial cost, defined as $f_{\text{plan}}:(I, Q, S, E) \rightarrow \tau$:
\begin{equation}
\min_{\tau} \sum_{i=1}^{n} W(p_i)
\end{equation}
subject to $p_1 = S$ and $p_n = E$.
\subsection{Symbolized Data Generation}
\subsubsection{Controllable Symbolic Query Synthesis}
We initiate the pipeline by sampling a configuration tuple $\sigma = (\text{agent}, \text{task}, \mathcal{C}_{\text{start/end}}, \mathcal{C}_{\text{allowed}})$. The ground-truth symbolic vectors are  deterministically derived via KB, which ensures that the underlying constraint is axiomatically governed by the rules defined in \cref{sec:kb_cons}.
\begin{equation}
f_{\text{KB}}(\sigma) \rightarrow (\mathbf{V}_{\text{trav}}^{gt}, \mathbf{V}_{\text{pref}}^{gt})
\end{equation}

We utilize DeepSeek-V3.2~\cite{liu2025deepseek} to synthesize the natural language query $Q$ based on $\sigma$. To enforce semantic alignment, the model is required to perform self-inference, re-deriving the logic vectors from its own generated text:
\begin{equation}
Q = \mathcal{M}(\sigma), \quad (\hat{\mathbf{V}}_{\text{trav}}, \hat{\mathbf{V}}_{\text{pref}}) = \mathcal{P}_{\text{self}}(Q)
\end{equation}
where $\mathcal{M}$ denotes the generative mapping and $\mathcal{P}_{\text{self}}$ represents the self-inference process, which ensures self-inferred vectors match the KB's GT. In addition, we utilize Gemini-3-Pro~\cite{gemini3pro2025} to verify that the textual descriptions in $Q$ for all $L=8$ land-cover classes are strictly compliant with the formal KB definitions, minimizing hallucinations and ensuring high-fidelity alignment between textual reasoning and symbolic constraints.

\subsubsection{Semantic Visual Grounding and Region Identification}
We ground the symbolic constraints from task 1 into segmentation masks provided by OpenEarthMap~\cite{xia2023openearthmap}.To maintain the structural integrity of critical terrains, we perform selective morphological filtering. Small isolated regions are filled with the majority class of their neighborhood, with the exception of functionally narrow classes such as Water (ID: 5), Road (ID: 3), and Building (ID: 7). This preserves the topological connectivity required for path planning while reducing visual clutter.
To ensure that each identified region provides a pure visual signal for Task 2, we apply iterative morphological erosion to large-scale land cover patches. Formally, for a region $R_i$ of class $c_k$:
\begin{equation}
\mathcal{E}(R_i) = \{ p \in R_i \mid \forall p' \in \mathcal{N}(p), c(p') = c_k \}
\end{equation}
where $\mathcal{N}(p)$ denotes the local neighborhood, ensuring the probe area is a maximum inscribed pure region. Instead of random pairing, we ensure that the visual scene $I$ can fully support the logic in $Q$. A query $Q$ is matched with an image $I$ only if the set of land covers present in the image $\mathcal{C}_{I}$ is a superset of the allowed and start/end classes specified in the query:$\mathcal{C}_{I} \supseteq \mathcal{C}_{\text{allowed}}$.
Specifically, we select the maximum area of each class present in $I$ as the primary region of interest, which forces the model to reason across diverse ground conditions and complex topological structures.

\subsubsection{Constrained Trajectory Generation and Optimization}
To ensure the feasibility of the planning mission under specific agent constraints, we first construct a region-level connectivity graph $\mathcal{G} = (V, E)$. Each node $v_i \in V$ represents a unique land-cover region $R_i$ identified in Stage 2. We construct an initial adjacency matrix $\mathbf{A} \in \{0, 1\}^{N \times N}$, where $A_{ij}=1$ if regions $R_i$ and $R_j$ are connected, and $0$ otherwise. To incorporate the symbolic constraints, we derive a constrained adjacency matrix $\tilde{\mathbf{A}}$ by masking $\mathbf{A}$ with $\mathbf{V}^{gt}_{\text{trav}}$:
\begin{equation}
\tilde{A}_{ij} = A_{ij} \cdot V^{gt}_{\text{trav}}[c(R_i)] \cdot V^{gt}_{\text{trav}}[c(R_j)]
\end{equation}
where $c(R_i)$ denotes the class ID of region $R_i$. A candidate pair of $(R_s,R_e)$ is validated if a path exists between them. To ensure a non-trivial navigation challenge and mitigate spatial biases, we specifically prioritize pairs $(R_s, R_e)$ that are spatially distant yet topologically connected.

Upon confirming reachability, we project the symbolic vectors onto the pixel grid to construct the cost map $W(p)$ following \cref{eq:costmap}. We then implement the A-Star search algorithm~\cite{hart1968formal} to find the optimal path. For any pixel $p$, the total estimated cost $f(p)$ is defined as:
\begin{equation}
f(p) = g(p) + h(p)
\end{equation}
where $g(p)$ is the actual cumulative cost from the start $S$ to $p$, and $h(p)$ is the heuristic function estimating the cost from $p$ to the destination $E$. We employ the Euclidean distance as the heuristic $h(p) = \|p - E\|_2$. The optimality of the generated path is guaranteed by the following properties:
\begin{itemize}
    \item Admissibility: Since the minimum cost for any traversable pixel is $W(p) \ge 1$ (following our cost definition), the Euclidean distance always satisfies $h(p) \le d^*(p, E)$, where $d^*$ is the actual minimum cost. 
    \item Consistency: Given that $W(p) \ge 1$ and the Euclidean distance satisfies the triangle inequality, the heuristic is consistent, meaning $h(p) \le W(p, q) + h(q)$ for any adjacent pixels $p$ and $q$.
\end{itemize}
Because $h(p)$ is both admissible and consistent, the A-Star algorithm is guaranteed to find the optimal trajectory $\tau^*$.

%% file: sections/4.Benchmark.tex
\section{NeSy-Route Benchmark}

\subsection{Dataset Statistics and Difficulty Stratification}
The NeSy-Route benchmark provides a large-scale collection of hierarchical Q-A pairs. Task 1 comprises 3,607 symbolic samples derived from the knowledge base, specifically designed to evaluate the model's comprehension of textual constraints. For Task 2, the reasoning complexity escalates with the semantic density of the visual scene; thus, we stratify the dataset into three difficulty tiers based on the number of annotated land-cover classes $M$ present in the image. This subset is partitioned into Easy ($1 \le M \le 4$) with 7,659 samples, Medium ($5 \le M \le 6$) with 3,712 samples, and Hard ($7 \le M \le L$) with 1,604 samples. We quantify the route planning challenge in Task 3 using Complexity Score ($\mathcal{D}$):
\begin{equation}
\mathcal{D} = \lambda_1 H_{\text{inter}} + \lambda_2 H_\text{intra} + \lambda_3 H_{\text{count}} + \lambda_4 \mathcal{C}_{\text{topo}} \label{eq:total_difficulty}
\end{equation}
where $H_{\text{inter}} = -\sum u_k \log_2 u_k$ and $H_{\text{count}} = -\sum v_k \log_2 v_k$ measure diversity via area proportions $u_k$ and region counts $v_k$; $H_{\text{intra}} = \frac{1}{L} \sum_k \frac{-\sum_j p_{kj} \log_2 p_{kj}}{\log_2 N_k}$ quantifies intra-class fragmentation with $p_{kj} = a_{kj}/A_k$ and region count $N_k$; and $\mathcal{C}_{\text{topo}} = \min ( 1.0, \frac{2 |E|}{|V| (|V| - 1)} )$ evaluates adjacency graph density. Based on $\mathcal{D}$, Task 3 is stratified into Easy (6,492 samples), Medium (2,705 samples), and Hard (1,624 samples), ensuring a granular assessment across complex landscapes.

% 定义颜色，避免过于刺眼，保持学术质感
\definecolor{DeepGreen}{RGB}{0, 150, 0}
\definecolor{DeepRed}{RGB}{200, 0, 0}

% 定义快捷命令，方便在表格中使用
\newcommand{\cmark}{\textcolor{DeepGreen}{\ding{51}}} % 粗体勾
\newcommand{\xmark}{\textcolor{DeepRed}{\ding{55}}}    % 粗体叉
\newcommand{\numc}[1]{\makebox[3.5em][c]{#1}} 
\begin{table}[t!]
\centering
\caption{Comparison of NeSy-Route with existing remote sensing benchmarks.}
\label{tab:final_comparison_full}
\resizebox{\columnwidth}{!}{ % 自动缩放以适应单栏宽度
\begin{tabular}{lcccccc}
\toprule
\multirow{2}{*}{Dataset} & \multicolumn{2}{c}{Key Statistic} & \multicolumn{2}{c}{Evaluation Paradigm} & \multicolumn{2}{c}{Construction Pipeline} \\ 
\cmidrule(lr){2-3} \cmidrule(lr){4-5} \cmidrule(lr){6-7}
&  Volume & Planning & Hierarchical & Symbolic & GT Optimality & Automated  \\ \midrule
% 定义一个固定宽度并居中的盒子，宽度约等于 "10,821" 的长度

RSVQA \cite{lobry2020rsvqa}       & - & \xmark & \cmark & \xmark & \cmark & \cmark \\
RSIVQA \cite{zheng2021mutual}     & - & \xmark & \xmark & \xmark & \cmark & \cmark \\
EarthVQA \cite{wang2024earthvqa}  & - & \xmark & \xmark & \xmark & \cmark & \xmark \\
LRSVQA \cite{luo2025large}        & - & \xmark & \cmark & \xmark & \xmark & \xmark \\
VRS-Bench \cite{li2024vrsbench}   & - & \xmark & \xmark & \xmark & \xmark & \xmark \\
FIT-RSRC \cite{luo2406skysensegpt} & - & \xmark & \cmark & \xmark & \xmark & \xmark \\
XLRS-Bench \cite{wang2025xlrs}    & 1,130 & \cmark & \cmark & \xmark & \cmark & \xmark \\ \midrule
NeSy-Route                        & \textbf{10,821}& \cmark & \cmark & \cmark & \cmark & \cmark \\ \bottomrule
\end{tabular}
}
\end{table}

\subsection{Symbolized Evaluator}
A typical example in NeSy-Route is shown in ~\cref{fig:behcmark}. To quantitatively evaluate the performance across three tasks, we employ a comprehensive set of metrics to assess the model's textual constraint understanding, text-image constraint alignment, and the feasibility and optimality of the generated trajectories.
\subsubsection{Task 1: Textual Constraint Understanding} 
For Task 1, we employ three primary metrics to assess the precision of symbolic constraint extraction from the query Q. First, the Traversability Matching (TM) measures the exact alignment of the predicted binary traversability vector with the ground truth:
\begin{equation} \label{eq:tm_metric}
TM = (1/N) \sum_{k=1}^{N} \mathbb{I}(\mathbf{V}_{\text{trav}, k} = \mathbf{V}^{gt}_{\text{trav}, k})
\end{equation}
where $\mathbb{I}$ is the indicator function and N is the total number of samples. Second, the Preference Ranking Correlation (PR) assesses the ordinal alignment of the preference vector. Let $M = \|\mathbf{V}^{gt}_{\text{trav}}\|_1$ denote the number of land cover classes permissible for navigation. We adopt the Kendall Tau coefficient \cite{kendall1938new} as the primary indicator, whcih has been used in many tasks~\cite{he2026monotonic, huang2024correlating, hamed2009effect}:
\begin{equation} \label{eq:pr_metric}
PR = (N_c - N_d) / \binom{M}{2}
\end{equation}
where $N_c$ and $N_d$ are scalar counts representing the concordant and discordant pairs among the $\binom{M}{2}$ possible combinations. Note that PR is strictly evaluated only for terrains where $V^{gt}_{\text{trav}}[i] = 1$ to ensure the ranking reflects the optimization logic among traversable surfaces. Finally, the Fully Matching Accuracy (FM) requires simultaneous mastery of both  traversability rules and preference rankings.:
\begin{equation} \label{eq:fm_metric}
FM = (1/N) \sum_{k=1}^{N} \mathbb{I}((\mathbf{V}_{\text{trav}, k} = \mathbf{V}^{gt}_{\text{trav}, k}) \wedge (\mathbf{V}_{\text{pref}, k} = \mathbf{V}^{gt}_{\text{pref}, k}))
\end{equation}
This allows for a granular analysis of how models internalize the symbolic logic embedded in textual constraints.

\subsubsection{Task 2: Text–Image Constraint Alignment}
For Task 2, we evaluate the text-image constraint alignment using the Region Matching Rate (RM), TM, and PR. The RM measures the global accuracy of region identification:
\begin{equation} \label{eq:rm_metric}
RM = (1 / (N \cdot L)) \sum_{k=1}^{N} \sum_{i=0}^{L-1} \mathbb{I}(\mathbf{V}'_{\text{reg}, k}[i] = \mathbf{V}^{gt}_{\text{reg}, k}[i])
\end{equation}
The TM and PR metrics are applied to the localized vectors $\mathbf{V}'_{\text{trav}}$ and $\mathbf{V}'_{\text{pref}}$ following \cref{eq:tm_metric} and \cref{eq:pr_metric}. This allows for the precise identification of failure bottlenecks within the cross-modal reasoning process.

\subsubsection{Task 3: Constrained Route Planning}
For Task 3, we evaluate the global trajectory planning by reconstructing a continuous dense path $\hat{\tau}$ from the predicted sparse waypoints $\tau$. Let $S_A = \{k \in \{1,\dots,N\} \mid \forall p \in \tau_k, W(p) < \infty\}$ denote the subset of samples where all waypoints reside in traversable regions, and $S_B = \{1,\dots,N\} \setminus S_A$ be the complementary subset of non-compliant samples. The Adherence Rate (AR) is thus defined as:
\begin{equation} \label{eq:ar_metric}
AR = |S_A| / N
\end{equation}
For samples in $S_A$, the dense trajectory $\hat{\tau}$ is reconstructed by connecting adjacent waypoints using the A-Star search algorithm based on the cost map $W(p)$. For samples in $S_B$, we utilize the Bresenham line algorithm \cite{bresenham1998algorithm} for interpolation. The Cost Ratio (CR) is calculated for $S_A$ to assess the optimality relative to $\tau^*$:
\begin{equation} \label{eq:cr_metric}
CR = (1/|S_A|) \sum_{k \in S_A} (\sum_{p \in \hat{\tau}_k} W(p) / \sum_{q \in \tau^*_k} W(q))
\end{equation}
Conversely, for the non-compliant samples in $S_B$, the Violation Ratio (VR) quantifies the proportion of pixels in the reconstructed path that infringe upon non-traversable regions:
\begin{equation} \label{eq:vr_metric}
VR = (1/|S_B|) \sum_{k \in S_B} (|\{p \in \hat{\tau}_k \mid W(p) = \infty\}| / |\hat{\tau}_k|)
\end{equation}
Finally, to evaluate the geometric proximity between the reconstructed dense trajectory $\hat{\tau}$ and the optimal trajectory $\tau^*$, we utilize the Chamfer Distance \cite{barrow1977parametric} (CD), a standard metric widely adopted in autonomous driving~\cite{shi2023error, zhang2023copilot4d, palladin2025self} for trajectory comparison, across all $N$ samples:
\begin{equation} \label{eq:cd_metric}
CD(\tau, \tau^*) = (1/N) \sum_{k=1}^{N} [(1/|\tau_k|) \sum_{p \in \hat{\tau_k}} \min_{q \in \tau^*_k} \|p-q\|_2 + (1/|\tau^*_k|) \sum_{q \in \tau^*_k} \min_{p \in \tau_k} \|q-p\|_2]
\end{equation}
This multi-dimensional metrics system enables a rigorous quantification of both logical adherence and spatial optimality in trajectory evaluation.

\subsection{Comparison with Existing Benchmarks}
Leveraging the comparative analysis in \cref{tab:final_comparison_full}, NeSy-Route distinguishes itself through three key innovations. First, it significantly expands the scale of remote sensing planning tasks, achieving an order-of-magnitude increase over XLRS-Bench~\cite{wang2025xlrs} with 10,821 rigorously constrained samples. Second, its unique hierarchical and symbolic evaluation paradigm decouples perception, reasoning, and planning, enabling researchers to precisely trace failure bottlenecks to specific cognitive stages. Finally, by integrating high-fidelity semantic segmentation with heuristic search, our automated data generation framework ensures that every ground-truth trajectory represents a mathematically proven global optimum, establishing an objective and highly extensible gold standard for route planning.

%% file: sections/5.Experiment.tex
\begin{table}[t!]
\centering
\caption{Performance comparison on Task 1: Textual Constraint Understanding. Metrics include Traversability Matching (TM), Preference Ranking Correlation (PR), and Fully Matching Accuracy (FM). The best and second-best results are highlighted in bold and underlined, respectively.}
\label{tab:performance_table_task1}
\begin{tabular}{l p{2cm} p{2cm}p{2cm} }
\toprule
Method &  \textbf{TM}$\uparrow$ & \textbf{PR}$\uparrow$ & \textbf{FM}$\uparrow$  \\ \midrule
\multicolumn{4}{c}{\textbf{Close-source Models}} \\  \midrule
GPT-5.1                    & 77.02& \underline{0.959} & \underline{69.20}     \\ 
Gemini-3-Pro               & \textbf{98.34} & \textbf{0.962} & \textbf{92.24}  \\ 
Qwen3-VL-Plus                      & 72.97 &0.923  &66.23   \\ \midrule
\multicolumn{4}{c}{\textbf{Open-source Models}} \\ \midrule
LLaVA-OneVision                    & 26.68 &0.187  & 9.32    \\ 
InternVL-3.5-8B                    &  52.48& 0.141 &  21.18  \\ 
Qwen3-VL-8B               &53.01 &0.068  &27.67    \\ 
Qwen3-VL-30B-A3B          &66.20  & 0.752 &  44.44  \\ 
Qwen3-VL-32B            &69.75  &-0.616  &  31.69  \\ 
Qwen3-VL-235B-A22B        &72.91&0.872  & 52.98   \\ 
Qwen3.5-27B                       &\underline{80.32} & 0.906 &63.52  \\
% SkySenseGPT                          &24.31& 0.194 & 8.51   \\ 
\bottomrule
\end{tabular}
\end{table}

\section{Experiment}

\subsection{Experimental Setup}
To evaluate the reasoning and planning capabilities of MLLMs on the NeSy-Route benchmark, we evaluate a diverse suite of state-of-the-art models grouped into two primary categories: (a) proprietary frontier models representing the current performance ceiling, including GPT-5.1 \cite{singh2025openai}, Gemini-3-Pro \cite{gemini3pro2025}, and Qwen3-VL-Plus \cite{bai2025qwen3}; (b) open-source MLLMs spanning various parameter scales and architectures to investigate performance in route planning tasks, including the Qwen3-VL series (8B-Instruct, 30B-A3B-Instruct, 32B-Instruct, and 235B-A22B-Instruct) \cite{bai2025qwen3}, Qwen-3.5-27B \cite{qwen3.5}, InternVL-3.5-8B \cite{wang2025internvl3}, and LLaVA-OneVision \cite{LLaVA-OneVision-1.5}. 
% For a rigorous and fair comparison, all models are evaluated in a zero-shot setting using uniform prompts for each task, ensuring that the results reflect their inherent perception, reasoning and planning capabilities. 
For a rigorous and fair comparison, Task 1 is evaluated in a few-shot setting with fixed format demonstrations, whereas Tasks 2 and 3 are evaluated in a zero-shot setting with only output-format instructions, ensuring that the results reflect models' inherent perception, reasoning, and planning capabilities. 
During the experiments we observe that both Geochat~\cite{kuckreja2024geochat} and SkySenseGPT~\cite{Guo_2024_CVPR_skysense} are built upon the LLaVA-1.5~\cite{liu2024improved} architecture. Neither model was trained with data requiring individual coordinate points as outputs, and both were designed for specific remote sensing tasks. As a result, their instruction-following capability is extremely limited, making them unable to complete the evaluation process of NeSy-Route. Therefore, the performance of these domain-specific models is not reported in our experimental results.

\begin{table}[t!]
\centering
\caption{Evaluation results for Task 2: Text–Image Constraint Alignment. The model performance is assessed via Region Matching Rate (RM), Traversability Matching (TM), and Preference Ranking Correlation (PR). The best and second-best results are indicated in bold and underlined, respectively.}
\label{tab:performance_table_task2}
\resizebox{\textwidth}{!}{ % 自动缩放以适应页面宽度
\begin{tabular}{l ccc ccc ccc ccc}
\toprule
\multirow{2}{*}{\textbf{Method}} & \multicolumn{3}{c}{\textbf{Easy}} & \multicolumn{3}{c}{\textbf{Medium}} & \multicolumn{3}{c}{\textbf{Hard}} & \multicolumn{3}{c}{\textbf{Avg}}\\
\cmidrule(lr){2-4} \cmidrule(lr){5-7} \cmidrule(lr){8-10} \cmidrule(lr){11-13}
& \textbf{RM}$\uparrow$ & \textbf{TM}$\uparrow$ & \textbf{PR}$\uparrow$ & \textbf{RM}$\uparrow$ & \textbf{TM}$\uparrow$ & \textbf{PR}$\uparrow$& \textbf{RM}$\uparrow$ & \textbf{TM}$\uparrow$ & \textbf{PR}$\uparrow$& \textbf{RM}$\uparrow$ & \textbf{TM}$\uparrow$ & \textbf{PR}$\uparrow$\\ 
\midrule
\multicolumn{13}{c}{\textbf{Close-source Models}} \\
\midrule
GPT-5.1           & 69.79 & 66.37 & 0.737 & 48.90 & 22.84 & \underline{0.567} & 50.09 & 16.72 & 0.510 &56.26 &35.31 & 0.605 \\
Gemini-3-Pro      & \textbf{78.16} & 58.14 & 0.582 & \textbf{63.17} &10.78  & 0.402 & \textbf{71.70} & 13.59 & 0.404 &\textbf{71.01} &27.50 &0.463 \\
Qwen3-VL-Plus     & \underline{72.44} & \underline{69.50} & \textbf{0.778} & 46.58 & 25.85 & 0.582 & 53.56 & 16.67 & 0.541 &57.53 &37.34 &\underline{0.634} \\
\midrule
\multicolumn{13}{c}{\textbf{Open-source Models}} \\
\midrule
LLaVA-OneVision   & 24.75 & 24.27 & -0.001  & 16.41 & 9.93 & 0.064 & 12.84 & 9.36 & 0.157 &18.00 &14.52 &0.073 \\
InternVL-3.5-8B   & 57.99 & 56.53 & 0.011 & 35.48 & 16.35 & 0.052 & 32.69 & 7.56 & 0.117 &42.05 &26.81 &0.060 \\
Qwen3-VL-8B        & 33.21 & 41.89 & -0.107 & 16.88 & 11.97 & -0.024 & 15.95 & 13.07 & 0.086 &22.01 &22.31 &-0.015 \\
Qwen3-VL-30B-A3B    & 72.27 & 65.16 & 0.280 & 51.00 & \underline{40.55} & 0.365 & 48.66 & \underline{22.65} & \underline{0.617} & 57.31& \underline{42.79}&0.421 \\
Qwen3-VL-32B       & 67.31 & 62.62 & -0.190 & 37.48 & 22.15 & -0.176 & 39.26 & 21.04 & -0.154 &48.02 &35.27 &-0.173 \\
Qwen3-VL-235B-A22B & 64.25 & \textbf{71.38} & \underline{0.749} & 33.81 & \textbf{43.80} & \textbf{0.667} & 32.92 & \textbf{43.20} & \textbf{0.690} & 43.66&\textbf{52.79} &\textbf{0.702} \\
Qwen3.5-27B      & 71.80 & 63.95 & 0.576 & \underline{52.13} & 16.04 & 0.483 & \underline{57.74} & 19.48 &0.505 & \underline{60.56}&33.16 &0.521  \\
% SkySenseGPT       & 23.71 & 22.44 & 0.063  & 14.49 & 11.20 & 0.124 & 12.90 & 10.47 & 0.116 &17.03 &14.70 &0.101 \\
\bottomrule
\end{tabular}
}
\end{table}
\subsection{Main Results}
\subsubsection{Task 1 reveals a clear performance gradient across models in textual constraint understanding.}
The results summarized in \cref{tab:performance_table_task1} further illustrate this trend across the evaluated models. Close-source Models, exemplified by Gemini-3-Pro, demonstrate constraint understanding capability. This indicates that frontier models are highly capable of precisely parsing nested hard constraints for heterogeneous agents and constructing rigorous logical chains of priority. Concurrently, Qwen-3.5-27B stands out as a representative open-source model, attaining a TM of 80.32\%, even surpassing certain closed-source counterparts which validates the constraint understanding capabilities.

\subsubsection{Task 2 indicates that effective text–image constraint alignment remains challenging for current MLLMs and critically relies on strong textual reasoning capabilities.}
Compared with performance in Task 1, all models experience a precipitous decline in performance upon the introduction of visual features, as shown in \cref{tab:performance_table_task2}, which underscores that while MLLMs may internalize textual constraint, their perception ability remains remarkably deficient. Although Gemini-3-Pro demonstrated the highest level of visual identification with RM of 71.01\%, it lagged significantly behind Qwen3-VL-Plus. This misalignment between visual perception and constraint validates the necessity of decoupled evaluation in Task 2. Notably, the Mixture-of-Experts (MoE) based Qwen3-VL-235B-A22B exhibited exceptional performance in logical alignment, with an average PR of 0.702 and TM of 52.79\%, both substantially surpassing GPT-5.1. Furthermore, empirical data show that model with poor logic parsing in Task 1 like LLaVA-OneVision maintains extremely low performance in Task 2, confirming that textual constraint understanding is the cornerstone of text-image constraint alignment.

\begin{table}[t!]
\centering
\caption{Experimental results on Task 3: Constrained Route Planning. The trajectory quality is measured by Adherence Rate (AR), Violation Ratio (VR), Cost Ratio (CR), and Chamfer Distance(CD). The top two methods are denoted by bold and underlined formats, respectively.
}
\label{tab:performance_table_task3}
\resizebox{\textwidth}{!}{ % 自动缩放以适应页面宽度
\begin{tabular}{l cccc cccc cccc cccc}
\toprule
\multirow{2}{*}{\textbf{Method}} & \multicolumn{4}{c}{\textbf{Easy}} & \multicolumn{4}{c}{\textbf{Medium}} & \multicolumn{4}{c}{\textbf{Hard}} & \multicolumn{4}{c}{\textbf{Avg}}\\
\cmidrule(lr){2-5} \cmidrule(lr){6-9} \cmidrule(lr){10-13} \cmidrule(lr){14-17}
 & \textbf{AR}$\uparrow$ & \textbf{VR}$\downarrow$ & \textbf{CR}$\downarrow$ & \textbf{CD}$\downarrow$ & \textbf{AR}$\uparrow$ & \textbf{VR}$\downarrow$ & \textbf{CR}$\downarrow$ & \textbf{CD}$\downarrow$ & \textbf{AR}$\uparrow$ & \textbf{VR}$\downarrow$ & \textbf{CR}$\downarrow$ & \textbf{CD}$\downarrow$ &
 \textbf{AR}$\uparrow$ & \textbf{VR}$\downarrow$ & \textbf{CR}$\downarrow$ & \textbf{CD}$\downarrow$\\ 
\midrule
\multicolumn{17}{c}{\textbf{Close-source Models}} \\
\midrule
GPT-5.1           & 25.90 & \underline{31.70} & \underline{1.47}& 66.71  & 14.90 & \underline{35.8} & 2.71& 74.79  & 12.10 & 38.90 & 1.32& 82.12 &17.63 &35.47 &\underline{1.83} & 74.54 \\
Gemini-3-Pro      & 27.40 & \textbf{28.30} & \textbf{1.20}& \underline{58.09} & 15.40 & \textbf{32.70} & \textbf{1.25}& 70.17  & 13.30 & \textbf{35.90} & \underline{1.28}& 77.96 &18.70 & \textbf{32.30}& \textbf{1.24}& 68.74 \\
Qwen3-VL-Plus     & \underline{42.80} & 35.30 & 7.16& \textbf{53.78}  & \underline{25.80} & 37.80 & \underline{6.61}& \textbf{63.36}  & \underline{20.40} & 39.60 & 2.49& \textbf{68.60}&\underline{29.67} &37.57 &5.42 &\textbf{61.91}  \\
\midrule\multicolumn{17}{c}{\textbf{Open-source Models}} \\
\midrule
LLaVA-OneVision   & 37.64 & 39.47 & 12.42 & 103.76 & 21.11 & 38.24 & 9.24 & 122.91 & 16.26 & 37.54 & 13.63 & 160.41 & 25.00&38.42 & 35.29&129.03\\
InternVL-3.5-8B   &  34.80& 36.50 & 14.82& 109.88  & 19.40 & 38.70 & 15.06& 127.07  & 15.10 & 41.30 & 21.38& 146.55 &23.10 &38.83 &17.09 &127.83 \\
Qwen3-VL-8B        & 40.30 & 37.20 & 13.58& 170.86  & 24.20 & 39.20 & 17.68& 181.78  & 18.40 & 41.60 & 10.64& 232.09 & 27.63& 39.33& 13.97& 194.91\\
Qwen3-VL-30B-A3B   & 36.70 & 35.70 & 9.36& 95.35  & 21.30 & 38.50 & 8.36& 107.58  & 16.90 & 40.70 & 6.12& 130.98 &24.97 &38.30 & 7.95&111.30 \\
Qwen3-VL-32B       & \textbf{46.00} & 36.50 & 7.39& 62.60  & \textbf{28.80} & 38.40 & 14.86& 75.36  & \textbf{23.30} & 40.40 & 16.35& 75.60 &\textbf{32.70} &38.43 & 12.87&71.19 \\
Qwen3-VL-235B-A22B & 40.90 & 35.50 & 10.78& 58.44  & 24.00 & 37.30 & 10.14& \underline{64.91}  & 18.50 & 39.10 & 5.19& \underline{69.50} &27.80 &37.30 &8.70 &\underline{64.28} \\
Qwen3.5-27B      & 30.10 & 31.90 & 2.29& 67.81  & 16.50 & 35.70 & 6.96& 86.89  & 13.70 & \underline{38.30} & \textbf{1.26}& 184.52 &20.10 &\underline{35.30} &10.51 &113.07 \\
% SkySenseGPT       & 35.60 & 38.24 & 12.62& 105.44  & 20.03 & 38.64& 7.36 & 125.21  & 15.63 & 40.64 & 15.52& 172.31  &23.75 &39.17 &11.83 &134.32\\
\bottomrule
\end{tabular}
}
\end{table}

\subsubsection{Task 3 demonstrates that constrained route planning remains a major challenge for current MLLMs, revealing a clear gap between perceptual understanding and effective planning solutions.}
As shown in \cref{tab:performance_table_task3},  closed-source models such as Gemini-3-Pro exhibit a distinct pattern of high-precision planning. Although the overall adherence rate of 18.70\% for Gemini-3-Pro is not the highest due to rigorous logical alignment requirements, its performance on compliant paths demonstrates strong optimality. Specifically, the average CR of 1.24 and CD of 68.74 for this model are the best compared with the optimal trajectory. 
In contrast, open-source models represented by Qwen3-VL-32B achieve a leading adherence rate of 32.70\%, demonstrating robust awareness of local obstacle avoidance. However, the average CR for these models remains at an elevated level of 12.87. This disparity suggests that while open-source models are often successful at remaining within traversable regions, they lack coherent global strategies, resulting in highly redundant and inefficient trajectories. 
Furthermore, performance metrics deteriorate significantly as environmental complexity increases from easy to hard levels. For example, the adherence rate of GPT-5.1 decreases from 25.90\% to 12.10\% as the violation ratio increases. These results confirm that path planning is a high-level cognitive capability that extends beyond simple semantic perception and reasoning. Even with a foundation in terrain perception and reasoning, models still struggle to produce complex and efficient solutions for route planning.

\subsubsection{Hierarchical Analysis}
Correlation analysis across the three integrated tasks confirms that strong perception and reasoning abilities do not necessarily translate into planning ability. The experimental results demonstrate that comprehensive perception and reasoning abilities constitute a necessary but insufficient condition for the successful execution of complex planning tasks. A deficiency in textual constraint understanding during the first task leads to certain failure in subsequent stages, whereas strong performance in the first and second tasks does not ensure success in the third task. This disparity indicates that models must bridge a fundamental cognitive gap between perceptual attribute recognition and constrained planning.

\subsubsection{Discussions}
Our experiments reveal two limitations of current MLLMs: 1)They struggle to incorporate land-type constraints, likely because existing training data emphasizes object recognition while neglecting land-type textures and geological characteristics. 2)Current remote sensing MLLMs rely on outdated architectures, limiting their ability to handle complex reasoning and planning tasks, which necessitates the development of powerful remote sensing models based on more advanced architectures.

%% file: sections/6.Conclusion.tex
\section{Conclusion}
In this paper, we introduce NeSy-Route, a large-scale neuro-symbolic benchmark designed to evaluate constrained route planning in remote sensing. The benchmark assesses the perception, reasoning and planning capabilities of MLLMs simultaneously through three hierarchical stages comprising textual constraint understanding, the text to image constraint alignment Task, and the constrained route planning Task. Leveraging high-fidelity semantic segmentation masks and heuristic search algorithms, we construct an algorithmically derived optimal route for each sample, which serves as an objective reference for evaluating model performance. Experimental results indicate that while current MLLMs demonstrate competence in pure constraint understanding, they face significant bottlenecks in text-image preception and planning capabilities. 
% NeSy-Route introduces a structured evaluation framework for constraint-aware route planning in remote sensing and provides a new benchmark for studying how MLLMs integrate perception, reasoning, and planning in complex geographic environments. 
NeSy-Route focuses on 2D semantic route planning over remote-sensing imagery, leaving agentic planning workflows and real-world physical interaction to future work. We hope NeSy-Route can support further research on MLLMs that better integrate perception, reasoning, and planning in complex geographic environments.

%% file: sections/7.Acknowledgements.tex
\section*{Acknowledgements}
This research was supported by the Jiangsu Science Foundation (BG2024036 and BK20243012) and the Natural Science Foundation of China (624B2068, 62576162, and 62576174).
It was also supported by the Fundamental Research Funds for the Central Universities (022114380023).